\documentclass[10pt,twocolumn,letterpaper]{article}
\pdfoutput=1

\usepackage[pagenumbers]{cvpr} % To force page numbers, e.g. for an arXiv version

\definecolor{cvprblue}{rgb}{0.21,0.49,0.74}
\usepackage[pagebackref,breaklinks,colorlinks,allcolors=cvprblue]{hyperref}

\usepackage{orcidlink}
\usepackage{multirow}
\usepackage{booktabs}

\usepackage{color}
\usepackage{colortbl}

\usepackage{algorithm}
\usepackage{algpseudocode}

\usepackage{amsfonts}
\usepackage{bbm}
\usepackage{bm}
\usepackage{wrapfig,lipsum,booktabs}
\usepackage[utf8]{inputenc}
\usepackage{url}
\usepackage{booktabs}
\usepackage{amssymb}
\usepackage{bbding}
\usepackage{pifont}
\usepackage{wasysym}
\usepackage{utfsym}
\usepackage{amsmath}
\usepackage{amssymb}
\usepackage{bbm}
\usepackage{fontawesome}

\usepackage{ragged2e}
\usepackage{graphicx}

\definecolor{my_blue}{HTML}{d3eaf2}
\definecolor{Green}{rgb}{0.85882353, 0.90980392, 0.84705882}
\definecolor{rose}{rgb}{0.60392157, 0.53333333, 0.43921569}
\definecolor{dred}{rgb}{0.7254902, 0.09803922, 0.10588235}
\definecolor{VAIL_Green}{rgb}{0, .7, .0}

\def\eg{\emph{e.g}\onedot} 
 
\def\paperID{2424} % *** Enter the Paper ID here
\def\confName{CVPR}
\def\confYear{2025}

\title{Hyperbolic Category Discovery}

\author{Yuanpei Liu\textsuperscript{*} \qquad Zhenqi He\textsuperscript{*} \qquad Kai Han\textsuperscript{\textdagger}\\
Visual AI Lab, The University of Hong Kong \\
{\tt\small \{ypliu0,zhenqi\_he\}@connect.hku.hk \qquad  kaihanx@hku.hk}}

\begin{document}
\maketitle
\renewcommand{\thefootnote}{\fnsymbol{footnote}}
\footnotetext[1]{Equal contribution.}
\footnotetext[2]{Corresponding author.}

\begin{abstract}

Generalized Category Discovery (GCD) is an intriguing open-world problem that has garnered increasing attention. Given a dataset that includes both labelled and unlabelled images, GCD aims to categorize all images in the unlabelled subset, regardless of whether they belong to known or unknown classes. 
In GCD, the common practice typically involves applying a spherical projection operator at the end of the self-supervised pretrained backbone, operating within Euclidean or spherical space. 
However, both of these spaces have been shown to be suboptimal for encoding samples that possesses hierarchical structures. 
In contrast, hyperbolic space exhibits exponential volume growth relative to radius, making it inherently strong at capturing the hierarchical structure of samples from both seen and unseen categories. 
Therefore, we propose to tackle the category discovery challenge in the hyperbolic space. 
We introduce HypCD, a simple \underline{Hyp}erbolic framework for learning hierarchy-aware representations and classifiers for generalized \underline{C}ategory \underline{D}iscovery. 
HypCD first transforms the Euclidean embedding space of the backbone network into hyperbolic space, facilitating subsequent representation and classification learning by considering both hyperbolic distance and the angle between samples. This approach is particularly helpful for knowledge transfer from known to unknown categories in GCD. 
We thoroughly evaluate HypCD on public GCD benchmarks, by applying it to various baseline and state-of-the-art methods, consistently achieving significant improvements. Project page: \url{https://visual-ai.github.io/hypcd/}

\end{abstract}    
\section{Introduction}
\label{sec:intro}
Recently, category discovery -- initially explored as novel category discovery (NCD)~\cite{han2019learning} and subsequently extended to generalized category discovery (GCD)~\cite{vaze2022generalized} -- has emerged as an intriguing open-world problem, gaining increasing attention. 
GCD addresses the challenges posed by partially labelled datasets, where the unlabelled subset may contain instances from both seen and unseen classes. The goal is to leverage knowledge from labelled data to effectively categorize the unlabelled data. 
Based on the way to predict category index, existing GCD methods can be broadly classified into two types: non-parametric methods~\cite{vaze2022generalized,hao2023cipr,rastegar2023learn,RastegarECCV2024} and parametric methods~\cite{wen2023parametric,wang2024sptnet,liu2025debgcd}. Non-parametric methods predict category index based on feature clustering while parametric methods utilize a parametric classifier.

\begin{figure}[t]
    \centering
    \includegraphics[width=0.47\textwidth]{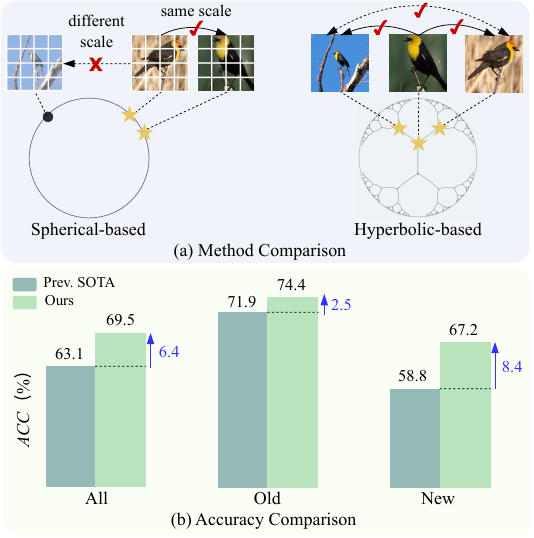}
    \caption{
    (a) Spherical-based \emph{vs.} Hyperbolic-based methods, where hyperbolic space better accommodates variations in scale and improves connections between samples. 
    (b) Average \textit{ACC} comparison of our method and previous SOTA across `All', `Old', and `New' categories on the SSB~\cite{vaze2022semantic} benchmark using DINO~\cite{caron2021emerging}.
    }
    \label{fig:intro}
\end{figure}

\begin{figure}[t]
    \centering
    \includegraphics[width=0.47\textwidth]{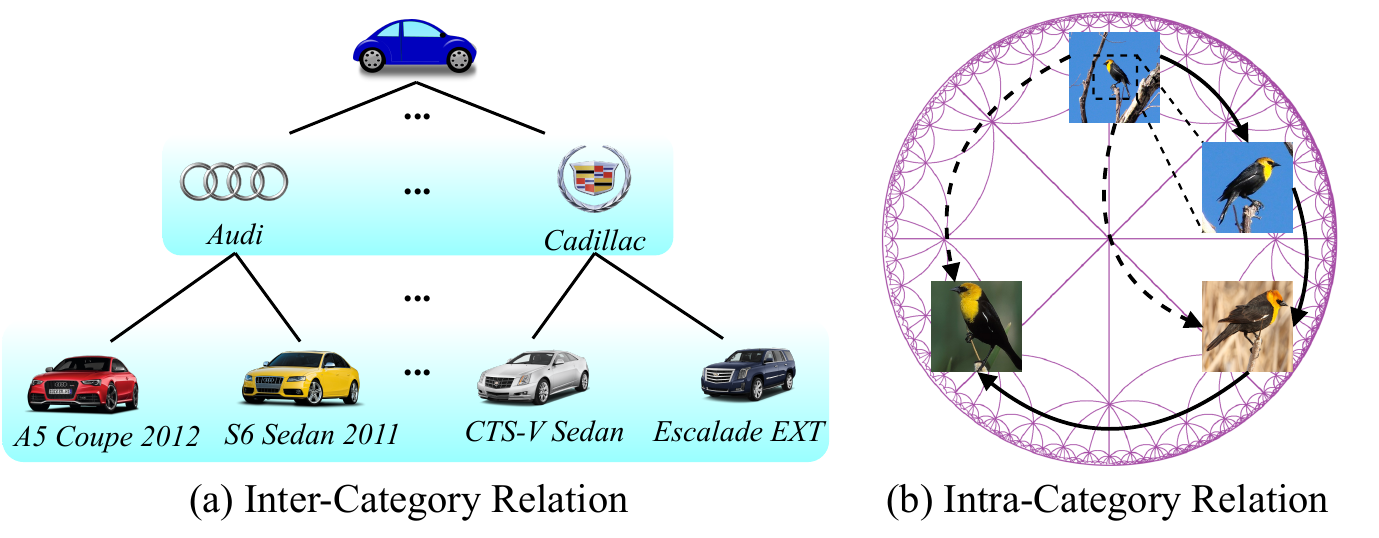}
    \caption{Hierarchical relations in GCD. (a) Inter-category relationships within the Stanford-Cars dataset~\cite{krause20133d}. (b) Intra-category relationships within CUB~\cite{wah2011caltech} dataset.}
    \label{fig:hrelations}
\end{figure}

As shown in the GCD literature~\cite{vaze2022generalized}, object parts are effective for knowledge transfer from seen to unseen categories, which is crucial for novel category discovery. 
Methods have been developed to explicitly learn better local features by learning pixel-level prompts around local image regions~\cite{wang2024sptnet} or utilizing part-level features~\cite{zhao2021novel}. 
However, these methods consider object parts as rigid image patches of the same size, without considering the hierarchical nature of the object parts and the scale discrepancy of the same parts in different images, thus unavoidably restricting the performance for GCD (see Fig.~\ref{fig:intro}(a)), in which the objects often have distinct poses, scales and appearance. 
To address this problem, one possible solution is to learn image embeddings possessing hierarchical constraints or following tree-like structures. 
This has been proven to be effective in many tasks. 
For example, in image retrieval and clustering, the hierarchy constraint may arise from whole-fragment relation~\cite{johnson2015image,Khrulkov_2020_CVPR}. 
Intuitively, in category discovery, which can be regarded as a \textit{transfer clustering} task~\cite{han2019learning}, we hypothesize that an embedding space that captures the hierarchical relations of object parts can also facilitate the discovery of new categories. 
Indeed, the hierarchical relations have been studied in GCD, such as~\cite{hao2023cipr,rastegar2023learn,wang2023discover,RastegarECCV2024,otholt2024guided}. However, they study from a substantially different perspective: inter-category hierarchy (see Fig.~\ref{fig:hrelations}(a)). This only considers the hierarchy of different semantic classes from \textit{coarse to fine} levels. 
Additionally, these methods require the relationships and number of levels in the hierarchy to be predefined, resulting in a lack of flexibility and scalability. 
Moreover, these methods are unable to capture more complex hierarchies, such as the compositional parts of an object (see Fig.~\ref{fig:hrelations}(b)). 
This is particularly the case because existing methods~\cite{vaze2022generalized,wen2023parametric,wang2024sptnet,RastegarECCV2024}, no matter whether they consider the hierarchy or not, learn the image embeddings in a spherical space. This follows the common practice of applying a spherical projection operator at the end of the self-supervised feature backbone~\cite{caron2021emerging,oquab2023dinov2}. 
Consequently, all subsequent operations, including distance calculations, are performed under either Euclidean or spherical geometry, resulting in limited awareness of hierarchical object parts.

In this work, we study the overlooked perspective in category discovery: \emph{instead of learning in the Euclidean or spherical space, we advocate a space that captures the hierarchical structure of each data point.} 
In spherical space, both the radius and volume are constant, whereas in Euclidean space, the volume grows polynomially with respect to the radius. 
Both of these spaces have been shown to be suboptimal for encoding samples that possess hierarchical structures~\cite{Khrulkov_2020_CVPR, ermolov2022hyperbolicvisiontransformerscombining, chen2022fullyhyperbolicneuralnetworks}. 
In contrast, hyperbolic space possesses a distinctive property where its volume grows exponentially relative to the radius. This characteristic makes hyperbolic space particularly suitable for embedding tree-like data, enhancing its representational power. 
Learning representations in hyperbolic space has proven to be effective in various computer vision tasks, including object recognition~\cite{ermolov2022hyperbolicvisiontransformerscombining}, object detection~\cite{kong2024hyperbolic}, semantic segmentation~\cite{weber2024flattening}, and anomaly detection~\cite{li2024hyperbolic}. 
Inspired by these successes, we aim to realize the idea of learning hierarchy-aware representations to facilitate knowledge transfer in category discovery, thereby unleashing the potential of hyperbolic representations.

To achieve this goal, we propose a simple yet effective framework, \textit{HypCD}, to properly learn the hierarchy-aware representation and classifier for category discovery through the lens of hyperbolic geometry. 
In this framework, we adapt our framework to popular  \textit{parametric}~\cite{wen2023parametric} and \textit{non-parametric}~\cite{vaze2022generalized} GCD baselines as well as the state-of-the-art (SOTA) method SelEx~\cite{RastegarECCV2024}, obtaining substantial improvements for them, establishing the new SOTA (see Fig.~\ref{fig:intro}(b)). 
\textit{Firstly}, starting from the self-supervised backbone pretrained in Euclidean space, we propose to map the Euclidean representation to a constrained Poincar\'{e} ball through feature clipping and exponential mapping. 
\textit{Secondly}, we implement the hyperbolic representation learning and build a hyperbolic classifier on the Poincar\'{e} ball, considering both angle and distance between samples in hyperbolic space. 
\textit{Thirdly}, to assign labels to unlabelled data after training, for non-parametric methods, we apply semi-supervised \textit{k-means} following the common practice; for parametric methods, we employ a hyperbolic classifier to make predictions. Despite its simplicity, our framework achieves significant performance improvements with two different pretrained weights (DINO~\cite{caron2021emerging} and DINOv2~\cite{oquab2023dinov2}) on the public GCD datasets, including the coarse-grained classification datasets CIFAR-10~\cite{krizhevsky2009learning}, CIFAR-100~\cite{krizhevsky2009learning}, and ImageNet-100~\cite{deng2009imagenet}, as well as the fine-grained SSB~\cite{vaze2021open} benchmark.

In summary, we make the following contributions in this paper:
\textbf{(i)} We identify the existing GCD methods' common shortcoming in encoding the hierarchical structure and propose to incorporate the hyperbolic geometry into the embedding space to address this limitation; 
\textbf{(ii)} We propose a simple yet effective framework, called \textit{HypCD}, for incorporating the hyperbolic geometry in representation learning and classification for category discovery; 
and \textbf{(iii)} Through extensive experiments on public GCD benchmarks by applying \textit{HypCD} to baseline and SOTA methods, our method consistently demonstrates effectiveness and superiority.

\section{Related Work}
\label{sec:related_work}

\paragraph{Category Discovery.}
{
Novel category discovery (NCD) is initially introduced in~\cite{han2019learning} to establish a realistic framework for transferring knowledge from seen categories to cluster unseen categories, by considering it as a transfer clustering problem. 
Many subsequent methods have been proposed to advance the field~\cite{han2020automatically, han2021autonovel, jia2021joint, zhao2021novel, ncl, fini2021unified}. 
Generalized category discovery~\cite{vaze2022generalized} (GCD) relaxes the assumption in NCD by considering unlabelled data containing samples from both known and unknown classes~\cite{vaze2022generalized}. 
Further investigations~\cite{cao2022open,hao2023cipr,pu2023dynamic,joseph2022novel,cendra2024promptccd,wang2024hilo,liu2025debgcd} have explored a variety of strategies to address the challenges posed by GCD. 
One notable approach, SimGCD~\cite{wen2023parametric}, proposes to learn a parametric classifier enhanced by mean entropy regularization, thereby improving performance. 
In another vein, GPC~\cite{Zhao_2023_ICCV} employs Gaussian mixture models to jointly learn robust representations while simultaneously estimating the number of unknown categories. 
SPTNet~\cite{wang2024sptnet} presents a spatial prompt tuning method that enables models to concentrate more effectively on specific object parts, thus enhancing knowledge transfer in GCD. 
Most recently, SelEx~\cite{RastegarECCV2024} has been proposed, leveraging hierarchical semi-supervised $k$-means to achieve SOTA results on fine-grained datasets. 
Additionally, various efforts are focused on addressing category discovery from multiple perspectives. For instance,~\cite{jia2021joint} emphasizes multi-modal category discovery; ~\cite{zhang2022grow} and~\cite{cendra2024promptccd} explore a continual setting; ~\cite{pu2024federated} studies category discovery in a federated setting; and  ~\cite{wang2024hilo} examines GCD in the presence of domain shifts.
}

\noindent\textbf{Hyperbolic Geometry.} 
Hyperbolic space, defined as a non-Euclidean manifold with exponential volume growth in relation to its radius, is inherently aligned with the embedding of tree-like and hierarchical data structures in visual recognition tasks. 
Significant advancements in this area include~\cite{Khrulkov_2020_CVPR}, which presents a hyperbolic image embedding technique by projecting model outputs into hyperbolic space, and~\cite{ermolov2022hyperbolicvisiontransformerscombining}, which integrates hyperbolic geometry into various vision transformer architectures, showcasing performance that surpasses their Euclidean counterparts. 
Hyperbolic methods have also been developed on diverse tasks such as image classification~\cite{Khrulkov_2020_CVPR, ermolov2022hyperbolicvisiontransformerscombining, guo2022clippedhyperbolicclassifierssuperhyperbolic}, action recognition~\cite{franco2023hyperbolic}, few-shot learning~\cite{hyperbolicFewshots2021iccv} and object segmentation~\cite{ghadimiatigh2022hyperbolicimagesegmentation, weng2021unsuperviseddiscoverylongtailinstance}. 
Moreover, recent developments have introduced hyperbolic geometry for neural networks including fully connected layers~\cite{shimizu2021hyperbolicneuralnetworks}, convolutional neural networks~\cite{bdeir2024fully}, graph neural networks~\cite{liu2019hyperbolicgraphneuralnetworks, chami2019hyperbolicgraphconvolutionalneural}, and attention network~\cite{gulcehre2018hyperbolicattentionnetworks}, thereby facilitating a deeper integration of hyperbolic geometry into deep learning regime. 

\section{Method}
In this section, we first introduce the task in Sec.~\ref{sec:method:task_setup}, then move to a review of baselines in Sec.~\ref{sec:method:baselines}. Afterwards, the geometry mapping and training details of our framework are described in Sec.~\ref{sec:method:hyp_space} and Sec.~\ref{sec:method:hyp_cd}. Lastly, the label assignment details are outlined in Sec.~\ref{sec:method:label_ass}.

\subsection{Problem Statement}
\label{sec:method:task_setup}
GCD aims to learn a model capable of accurately classifying unlabelled samples from known categories while simultaneously clustering those from unknown categories. 
Consider an unlabelled dataset denoted as $\mathbf{D}_u = \{(\mathbf{x}^{u}_{i}, {y}^{u}_{i})\} \in \mathbf{X} \times \mathbf{Y}_u$ and a labelled dataset represented as $\mathbf{D}_l = \{(\mathbf{x}^{l}_{i}, {y}^{l}_{i})\} \in \mathbf{X} \times \mathbf{Y}_l$, where $\mathbf{Y}_u$ and $\mathbf{Y}_l$ denote the respective label sets. 
The unlabelled dataset comprises samples from both known and unknown categories, \textit{i.e.}, specifically $\mathbf{Y}_l \subset \mathbf{Y}_u$. 
Let the number of labelled categories be denoted by $M=|\mathbf{Y}_l|$. 
We assume that the total number of categories, $K=|\mathbf{Y}_l \cup \mathbf{Y}_u|$, is known, as posited in prior works~\cite{han2021autonovel,wen2023parametric,vaze2023no}. 
In scenarios where this information is unavailable, alternative methods such as those proposed in~\cite{han2019learning,vaze2022generalized} can be employed to yield a reliable estimation.

\subsection{Review of Baselines}
\label{sec:method:baselines}
\noindent\textbf{Non-parametric Baseline.}
\cite{vaze2022generalized} formalizes the GCD task and proposes a non-parametric baseline. The approach involves finetuning the pre-trained DINO~\cite{caron2021emerging} model~\cite{dosovitskiy2020image} to enhance the learned representation. 
The loss function comprises a supervised contrastive loss, which operates on the labelled samples, and a self-supervised contrastive loss, which operates on all the samples. 
Specifically, given two randomly augmented views $\mathbf{x}_i$ and $\mathbf{x}_i'$ for the same image in a mini-batch ${B}$, the self-supervised contrastive loss is:
\begin{equation}
\textstyle
    \mathcal{L}_{rep}^{u} = \frac{1}{|{B}|}\sum_{i\in {B}} - \text{log}\frac{\exp(\mathbf{z}_i \cdot \mathbf{z}_i'/\tau_r)}{\sum\nolimits_j^{j\neq i}\exp(\mathbf{z}_i\cdot \mathbf{z}_j'/\tau_r)},
\end{equation}
where the feature $\mathbf{z}_i=\rho_r(\phi(\mathbf{x}_i))$ is a $\ell_2$-normalized vector and $\mathbf{z}_i'$ represents feature from another view $\mathbf{x}_i'$. Here, $\phi$ refers to the backbone network, $\rho_r$ denotes the projection head, and $\tau_r$ stands as the temperature parameter used for scaling the features. 
The supervised contrastive loss for labelled samples is:
\begin{equation}
\textstyle
    \mathcal{L}_{rep}^{s} = \frac{1}{|{B}_l|}\sum_{i\in {B}_l} \frac{1}{|{N}_i|}\sum\limits_{q\in {N}_i}-\text{log}\frac{\exp(\mathbf{z}_i \cdot \mathbf{z}_q/\tau_r)}{\sum\nolimits_j^{j\neq i}\exp(\mathbf{z}_i \cdot \mathbf{z}_j/\tau_r)},
\end{equation}
where ${N}_i$ is the index set for all other images in the labelled mini-batch ${B}_l \subset {B}$ having the same label as $\mathbf{x}_i$. The overall representation learning loss is then: $\mathcal{L}_{rep}=(1-\lambda_b)\mathcal{L}_{rep}^{u} + \lambda_b\mathcal{L}_{rep}^{s}$, where $\lambda_b$ is a balance factor. 

\noindent\textbf{Parametric Baseline.}
\label{Sec:Parametric}
\cite{wen2023parametric} introduces a robust parametric GCD baseline, which has been widely adopted in the field ever since~\cite{vaze2023no,wang2024sptnet}. This method employs a parametric classifier implemented in a self-distillation framework~\cite{caron2021emerging}. 
The classifier is randomly initialized with $K$ normalized category prototypes $\mathbf{C}=\{\mathbf{c}_1,...,\mathbf{c}_K\}$. For a randomly augmented view $\mathbf{x}_i$ and its corresponding normalized hidden feature vector $\mathbf{h}_i=\phi(\mathbf{x}_i)/||\phi(\mathbf{x}_i)||$, the output probability for the $k$-th category is given by: 
\begin{equation}
\textstyle
    {\mathbf{p}_i}^{(k)} = \frac{\exp(\mathbf{h}_i\cdot\mathbf{c}_k/\tau_s)}{\sum\nolimits_{j=1}^K \exp(\mathbf{h}_i\cdot\mathbf{c}_j/\tau_s)},
\end{equation}
where $\tau_s$ is the scaling temperature for the \textit{`student'} view. The soft label $\mathbf{q}_i$ is generated by the \textit{`teacher'} view with a sharper temperature $\tau_t$ based on another augmented view in a similar manner.  
The self-distillation loss for the two views is then computed using the cross-entropy loss function: $\ell_{ce}(\mathbf{q}',\mathbf{p})=-\sum\nolimits_{j=1}^K \mathbf{q}'^{(j)}\text{log}~\mathbf{p}^{(j)}$. 
The unsupervised loss is then computed by aggregating contributions from all samples in the mini-batch $B$ as follows:
\begin{equation}
\textstyle
    \mathcal{L}_{cls}^{u}=\frac{1}{|{B}|}\sum_{i\in {B}} \ell_{ce}(\mathbf{q}'_i,\mathbf{p}_i)-\xi \mathcal{H}(\overline{\mathbf{p}}), 
\end{equation}
where $\overline{\mathbf{p}}=\frac{1}{2|{B}|}\sum\nolimits_{i\in {B}}(\mathbf{p}_i+\mathbf{p}'_i)$ denotes the mean prediction across the mini-batch. The mean entropy is defined as: $\mathcal{H}(\overline{\mathbf{p}})=-\sum\nolimits_{j=1}^K \overline{\mathbf{p}}^{(j)}\text{log}~\overline{\mathbf{p}}^{(j)}$, weighted by $\xi$. 

For the labelled samples, the supervised classification loss is written as $\mathcal{L}_{cls}^s=\frac{1}{|{B}_l|}\sum\nolimits_{i\in {B}_l}\ell_{ce}(\mathbf{p}_i,\mathbf{y}_i)$, where $\mathbf{y}_i$ represents the one-hot vector corresponding to the ground-truth label $y_i$. 
The overall objective is $\mathcal{L}_{cls}=(1-\lambda_b)\mathcal{L}_{cls}^u+\lambda_b\mathcal{L}_{cls}^s$. Integrating this with the representation learning loss $\mathcal{L}_{rep}$ adopted from~\cite{vaze2022generalized}, the comprehensive training objective is expressed as: $\mathcal{L}_{gcd} = \mathcal{L}_{cls} + \mathcal{L}_{rep}.$ 
Through training with $\mathcal{L}_{gcd}$ on both $\mathbf{D}_l$ and $\mathbf{D}_u$, the classifier is empowered to directly predict labels for the unlabelled samples after the training process concludes. 

\begin{figure*}
    \centering
    \includegraphics[width=1.0\textwidth]{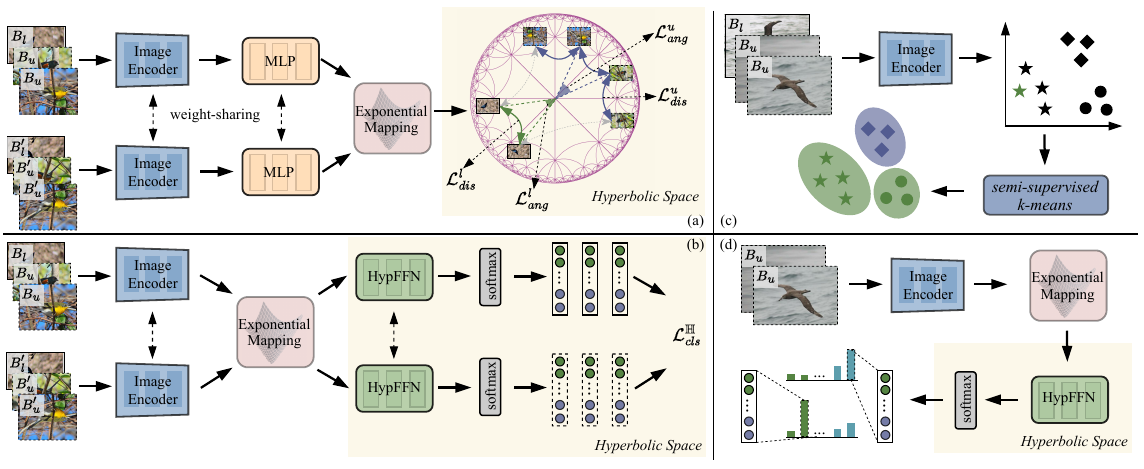}
    \caption{Overall pipeline of our HypCD framework for parametric and non-parametric GCD baselines. (a) Hyperbolic representation learning. (b) Hyperbolic classifier. (c) Non-parametric label assignment. (d) Parametric label assignment.
    }
    \label{fig:method}
\end{figure*}

\subsection{Hyperbolic Space for Category Discovery}
\label{sec:method:hyp_space}
As previously discussed, object parts are critical for facilitating knowledge transfer from labelled categories to unseen ones in GCD. Each sample inherently contains object parts that reside within a hierarchical structure. 
Moreover, existing GCD methods~\cite{rastegar2023learn,RastegarECCV2024} emphasize the inter-category hierarchy to enhance the clustering performance of unlabelled samples in Euclidean or spherical spaces. 
However, the geometry of representation space limits their ability to effectively capture other kinds of hierarchy~\cite{Khrulkov_2020_CVPR}. 
In contrast, hyperbolic space, characterized by its property of exponential volume growth with respect to the radius~\cite{ermolov2022hyperbolicvisiontransformerscombining}, emerges as a more suitable space for GCD.

Hyperbolic space $\mathbb{H}^n$ is defined as an  $n$-dimensional Riemannian manifold exhibiting constant negative curvature, and it encompasses several analytic models~\cite{cannon1997hyperbolic}. 
Following previous literature~\cite{Khrulkov_2020_CVPR, ermolov2022hyperbolicvisiontransformerscombining}, we employ the \textit{Poincar\'{e} ball}~\cite{nickel2017poincare} model. 
In this model, the hyperbolic space is represented as an $n$-dimensional ball $\mathbb{D}_c^n = \left\{ \mathbf{a} \in \mathbb{R}^n \ \big| \ c\| \mathbf{a} \|^2 < 1 \right\}$ with \textit{curvature value} $-c^2$, where $c$ is the non-negative curvature parameter. 
The manifold is equipped with the Riemannian metric $g^\mathbb{D} = \lambda_c^2 \, g^\mathbb{E}$ where $\lambda_c (\mathbf{a}) = \frac{2}{1 - c\| \mathbf{a} \|^2}$ is the \textit{conformal factor} and $g^\mathbb{E}$ is the \textit{identity metric} $\mathbf{I}_n$ in Euclidean space. In this way, the local distances are scaled by the factor $\lambda_c$ approaching infinity near the boundary of the ball. This gives rise to the \textit{exponential expansion} property of hyperbolic spaces, unlike the {polynomial expansion} in Euclidean space. 
However, hyperbolic space is not vector space and thus operations such as addition can not be directly conducted. To address this problem, we leverage the \textit{gyrovector formalism}~\cite{Ungar2008}. For a pair of points $\mathbf{a},\mathbf{b} \in \mathbb{D}_c^n$, their \textit{M\"{o}bius addition} is defined as:

\begin{equation}
\textstyle
\label{equ:addition}
    \mathbf{a} \oplus_c \mathbf{b} = \frac{(1+2c\langle \mathbf{a}, \mathbf{b} \rangle + c\|\mathbf{b}\|^2) \mathbf{a}+ (1-c\|\mathbf{a}\|^2)\mathbf{b}}{1+2c\langle \mathbf{a}, \mathbf{b} \rangle + c^2 \|\mathbf{a}\|^2 \|\mathbf{b}\|^2}.
\end{equation}
The \textit{hyperbolic distance} between them is then: 
\begin{equation}\label{eq:hdist}
\textstyle
    \mathcal{D}_{\mathbb{H}}(\mathbf{a},\mathbf{b}) = \frac{2}{\sqrt{c}} \mathrm{arctanh}(\sqrt{c}\|-\mathbf{a} \oplus_c \mathbf{b}\|)
\end{equation}
When $c\rightarrow 0$, the hyperbolic distance (Eq.~\ref{eq:hdist}) closes to the Euclidean distance $\lim_{c \to 0} \mathcal{D}_{\mathbb{H}}(\mathbf{a},\mathbf{b})=2\|\mathbf{a}-\mathbf{b}\|.$

\subsection{HypCD}
\label{sec:method:hyp_cd}
As illustrated in Fig.~\ref{fig:method}, we propose a unified framework, HypCD, for category discovery in hyperbolic space, incorporating both parametric~\cite{wen2023parametric} and non-parametric GCD approaches. 
Given two randomly augmented views, we initially obtain the respective Euclidean feature vectors $\mathbf{z}_i$ and $\mathbf{z}_i'$ through a self-supervised pretrained backbone~\cite{caron2021emerging, oquab2023dinov2}. 
Subsequently, the feature embeddings are mapped into hyperbolic space $\mathbb{H}^n$ using \textit{exponential mapping}, facilitating representation learning within this exponentially growing space to more effectively capture and utilize the hierarchical relationships inherent in the training data. 

The exponential mapping~\cite{Khrulkov_2020_CVPR} serves as a bijective projection between Euclidean space $\mathbb{E}^n$ and hyperbolic space $\mathbb{H}^n$. The projection of \textit{tangent vector} $\mathbf{z}$ from $\mathbb{E}^n$ to $\mathbb{H}^n$ is formulated as:
\begin{equation}\label{eq:exp}
    \exp_\mathbf{\mathbf{o}}^c(\mathbf{z}) = \mathbf{o} \oplus_ c \bigg(\tanh \bigg(\sqrt{c} \frac{\lambda_\mathbf{o}^c \|\mathbf{z}\|}{2} \bigg) \frac{\mathbf{z}}{\sqrt{c}\|\mathbf{z}\|}\bigg),
\end{equation}
where $\oplus_c$ is the \textit{M\"{o}bius addition}, as introduced in Eq.~\ref{equ:addition} and $\mathbf{o}$ represents the \textit{base point} of the mapping. 
To address the issue of \textit{gradient vanishing}~\cite{guo2022clippedhyperbolicclassifierssuperhyperbolic} near the boundary of the Poincaré ball during training, we implement a \textit{feature clipping} operation in $\mathbb{E}^n$ prior to the exponential mapping. The operation is defined as: $\mathcal{C}(\mathbf{z}) = \min \{1, \frac{r}{||\mathbf{z}||}\}\cdot \mathbf{z}$, where $r$ denotes the clipping value. 
For the feature vector $\mathbf{z}_i$ in $\mathbb{E}^n$, the corresponding mapped feature in $\mathbb{H}^n$ is expressed as $\mathcal{M}(\mathbf{z}_i) = \exp_\mathbf{\mathbf{o}}^c(\mathcal{C}(\mathbf{z}_i))$. The same operation will also be applied to the other feature vector $\mathbf{z}_i'$.

As described in Sec.\ref{sec:method:baselines}, both parametric~\cite{vaze2022generalized} and non-parametric~\cite{wen2023parametric} baselines utilize the same representation learning method. In our framework, we implement a consistent representation learning solution in hyperbolic space for them (Fig.\ref{fig:method}(a)). For parametric approaches, a hyperbolic parametric classifier is employed (Fig.~\ref{fig:method}(b)). We will introduce these components in detail subsequently.

\noindent\textbf{Hyperbolic Representation Learning.}
Following prior attempts~\cite{vaze2022generalized, wen2023parametric, RastegarECCV2024}, we incorporate both self-supervised and supervised contrastive learning into our framework. 
However, our approach uniquely operates within hyperbolic space. 
Furthermore, unlike previous GCD methods that exclusively utilize cosine distance~\cite{vaze2022generalized, wen2023parametric, wang2024sptnet} (\textit{angle-based}) or Euclidean distance~\cite{RastegarECCV2024} (\textit{distance-based}) for calculating pairwise similarity, we propose a hybrid approach that combines both distance-based and angle-based losses. Such integration has been shown to be more effective for model optimization in hyperbolic space~\cite{franco2023hyperbolic}. 
First, the unified form of self-supervised contrastive loss can be defined as:
\begin{equation}
\textstyle
\mathcal{L}^u = \frac{1}{|{B}|}\sum\limits_{i\in {B}} - \text{log}\frac{\exp(\mathcal{S}(\mathcal{M}(\mathbf{z}_i), \mathcal{M}(\mathbf{z}_i'))/\tau_r)}{\sum\nolimits_j^{j\neq i}\exp(\mathcal{S}(\mathcal{M}(\mathbf{z}_i), \mathcal{M}(\mathbf{z}_j^\prime))/\tau_r)}.
\end{equation}
Similarly, the supervised contrastive loss is unified as: 
\begin{equation}
\textstyle
    \mathcal{L}^s=\frac{1}{|{B}_l|}\!\!\sum\limits_{i\in {B}_l}\!\!\frac{1}{|{N}_i|}\!\!\!\sum\limits_{q\in {N}_i}\!\!\!-\text{log}\frac{\exp(\mathcal{S}(\mathcal{M}(\mathbf{z}_i), \mathcal{M}(\mathbf{z}_q))/\tau_r)}{\sum\limits_{j\neq i}\exp(\mathcal{S}(\mathcal{M}(\mathbf{z}_i), \mathcal{M}(\mathbf{z}_j))/\tau_r)},
\end{equation}
where $\mathcal{S}$ denotes the similarity function, which can be either distance-based or angle-based. 
For distance-based contrastive loss, we utilize $\mathcal{S}_d = -\mathcal{D}_{\mathbb{H}}$ as the similarity function, which is formally computed using negative Euclidean distance in prior methods~\cite{RastegarECCV2024}. 
For angle-based contrastive loss, we employ the \textit{cosine similarity}, formulated as: 
\begin{equation}
\textstyle
\mathcal{S}_a(\mathcal{M}(\mathbf{z}_i), \mathcal{M}(\mathbf{z}_i')) =\frac{\mathcal{M}(\mathbf{z}_i) \cdot \mathcal{M}(\mathbf{z}_i')}{||\mathcal{M}(\mathbf{z}_i)|| \cdot ||\mathcal{M}(\mathbf{z}_i')||}.
\end{equation}
Since hyperbolic space is \textit{conformal} with Euclidean space, cosine similarity remains equivalent in both $\mathbb{E}^n$ and $\mathbb{H}^n$.

The final supervised and self-supervised hyperbolic contrastive loss is composed of both distance-based and angle-based losses: 
\begin{equation}
\textstyle
\begin{split}
    &  \mathcal{L}_{hrep}^{s} = \alpha_{d}\mathcal{L}_{dis}^{s} + (1-\alpha_{d})\mathcal{L}_{ang}^{s}, \\
    &  \mathcal{L}_{hrep}^{u} = \alpha_{d}\mathcal{L}_{dis}^{u} + (1-\alpha_{d})\mathcal{L}_{ang}^{u},
\end{split}
\end{equation}
where $\mathcal{L}_{hrep}^{s} $ and $\mathcal{L}_{hrep}^{u}$ represent the supervised and self-supervised hyperbolic contrastive loss, respectively. The terms $\mathcal{L}_{dis}$ and $\mathcal{L}_{ang}$ correspond to distance-based and angle-based contrastive loss, respectively, obtained by substituting $\mathcal{S}$ with $\mathcal{S}_d$ and $\mathcal{S}_a$. Additionally, $\alpha_d$ is the loss weight of distance-based loss.  
The overall training objective for hyperbolic representation learning is: 
\begin{equation}
{
\textstyle
    \mathcal{L}_{rep}^\mathbb{H} = (1-\lambda_{b}^\mathbb{H})\mathcal{L}_{hrep}^{u} + \lambda_{b}^\mathbb{H} \mathcal{L}_{hrep}^{s},
}
\end{equation}
where $\lambda_{b}^\mathbb{H}$ serves as the balancing factor between the supervised and unsupervised losses.

\noindent\textbf{Hyperbolic Classifier.}
To enhance the parametric baseline with hyperbolic geometry, we replace the conventional Euclidean classification head—traditionally reliant on a multilayer perceptron (MLP) in Euclidean space—with its hyperbolic counterpart, the hyperbolic feed forward network (\texttt{HypFFN}). The \textit{hyperbolic linear} layer~\cite{ganea2018hyperbolic} exhibits greater alignment with the baseline~\cite{wen2023parametric}, and we experimentally find that it outperforms the \textit{hyperbolic multinomial logistic regression} layer. 
Consider the last linear layer of the MLP; similar to its Euclidean counterpart, the hyperbolic linear layer is parameterized by a weight matrix $\mathbf{w} \in \mathbb{R}^{I \times K}$ and a bias vector $\mathbf{s} \in \mathbb{R}^{1 \times K} $, where $I$ denotes the input feature dimension. 
Given the hyperbolic feature $\mathbf{z}^{\mathbb{H}}_i=\mathcal{M}(\mathbf{z}_i) \in \mathbb{R}^{1 \times I}$, the linear layer operates as \texttt{HypLinear}$(\mathbf{z}^{\mathbb{H}}_i, \mathbf{w}, \mathbf{s}) = \texttt{Proj} [(\mathbf{w} \otimes_c \mathbf{z}^{\mathbb{H}}_i) \oplus_c \mathbf{s}]$, where $\oplus_c$ follows Eq.~\ref{equ:addition}. The \textit{M\"{o}bius matrix-vector multiplication} $\mathbf{v}_i = \mathbf{w} \otimes_c \mathbf{z}^{\mathbb{H}}_i$ is defined as:
\begin{equation}
\textstyle
    \frac{1}{\sqrt{c}} \tanh\left(
            \frac{\|\mathbf{z}^{\mathbb{H}}_i \mathbf{w}\|_2}{\|\mathbf{z}^{\mathbb{H}}_i\|_2}\tanh^{-1}(\sqrt{c}\|\mathbf{z}^{\mathbb{H}}_i\|_2)
        \right)\frac{\mathbf{z}^{\mathbb{H}}_i\mathbf{w}}{\|\mathbf{z}^{\mathbb{H}}_i\mathbf{w}\|_2}.
\end{equation}
To ensure \textit{numerical stability}~\cite{ganea2018hyperbolic}, a safe projection is operated on the result manifold and represented as:
\begin{equation}
\textstyle
    \texttt{Proj} (\mathbf{v}_i) \!=\!\!\\
\begin{cases} 
\frac{\mathbf{v}_i}{\|\mathbf{v}_i\|_2}\times \frac{1-10^{-3}}{\sqrt{c}}, &\!\!\!\!\!\frac{1-10^{-3}}{\sqrt{c}}<\|\mathbf{v}_i\|_2  \\
\mathbf{v}_i, & \text{otherwise} 
\end{cases}.
\end{equation}
This integration allows our hyperbolic classifier to be seamlessly incorporated into the baseline~\cite{wen2023parametric} by substituting the original MLP with \texttt{HypFFN}. 
For each point in $\mathbb{H}^n$, the \textit{tangent space} at that point serves as a Euclidean subspace, enabling straightforward adaptation of Euclidean operations within this space~\cite{chen2022fullyhyperbolicneuralnetworks}. 
Consequently, the cross-entropy loss for the hyperbolic classifier can be expressed as: $\ell_{ce}^{\mathbb{H}} = \ell_{ce}(\texttt{HypFFN}(\mathbf{z}^{\mathbb{H}}_i), \mathbf{y}_i)$. 
Additionally, we can define the hyperbolic counterpart $\mathcal{H}^{\mathbb{H}}$ for the mean entropy $\mathcal{H}$. 
By substituting the original $\ell_{ce}$ and $\mathcal{H}$ with our derived $\ell_{ce}^{\mathbb{H}}$ and $\mathcal{H}^{\mathbb{H}}$, respectively, we can readily compute the final hyperbolic classifier loss $\mathcal{L}^{\mathbb{H}}_{cls}$, as detailed in Sec.~\ref{sec:method:baselines}. 

\subsection{Label Assignment}
\label{sec:method:label_ass}
Existing approaches typically employ either a parametric classification head or non-parametric methods, such as semi-supervised \textit{k-means}~\cite{vaze2022generalized}, for label assignment. In this paper, we do not independently assess these two methods; rather, we integrate both within the HypCD framework as shown in Fig.~\ref{fig:method}(c) and (d). 
For non-parametric approaches, including ~\cite{vaze2022generalized} and the recent SelEx~\cite{RastegarECCV2024}, we retain the original label assignment strategy by applying semi-supervised \textit{k-means} clustering directly to feature representations extracted by $\phi$ in $\mathbb{E}^n$. Our empirical results indicate that training in hyperbolic space allows for the transfer of hierarchical structure encoding from $\mathbb{H}^n$ to $\mathbb{E}^n$. Moreover, we find that the operations of \textit{k-means} in $\mathbb{E}^n$ are significantly more efficient while maintaining comparable performance. 
For the parametric baseline exemplified by SimGCD~\cite{wen2023parametric}, we utilize the hyperbolic classification head to conduct classification within hyperbolic space using the trained hyperbolic classifier \texttt{HypFFN}. 
Both design choices are theoretically supported by the property of hyperbolic geometry of encoding hierarchical structures, facilitating a more intuitive and effective representation and classifier for GCD.

\section{Experiment}
\label{sec:experiment}
\begin{table*}[ht]
\vspace{-0.5em}
  \caption{Comparison of GCD methods on the SSB~\cite{vaze2022semantic} benchmark, CIFAR-10~\cite{krizhevsky2009learning}, CIFAR-100~\cite{krizhevsky2009learning} and ImageNet-100~\cite{deng2009imagenet} datasets. Results are reported in \textit{ACC} across the `All', `Old' and `New' categories.}
  \label{tab:results}
  \centering
  \setlength{\tabcolsep}{1.5mm}{
  \resizebox{1.0\linewidth}{!}{
\begin{tabular}{@{}llcccccccccccccccccc@{}}
\toprule
&&\multicolumn{3}{c}{CUB~\cite{wah2011caltech}}& \multicolumn{3}{c}{Stanford-Cars~\cite{krause20133d}}&\multicolumn{3}{c}{FGVC-Aircraft~\cite{maji2013fine}}&\multicolumn{3}{c}{CIFAR-10~\cite{krizhevsky2009learning}}& \multicolumn{3}{c}{CIFAR-100~\cite{krizhevsky2009learning}}&\multicolumn{3}{c}{ImageNet-100~\cite{deng2009imagenet}}\\ \cmidrule(lr){3-5} \cmidrule(lr){6-8} \cmidrule(lr){9-11} \cmidrule(lr){12-14} \cmidrule(lr){15-17} \cmidrule(lr){18-20}&
Method
&All&Old&New&All&Old&New&All&Old&New&All&Old&New &All&Old&New &All&Old&New\\
\midrule
\multirow{21}{*}{\rotatebox{90}{\emph{DINO}}}
&ORCA~\cite{cao2022open}&36.3&43.8&32.6 &31.6&32.0&31.4 &31.9&42.2&26.9 &69.0&77.4&52.0 &73.5&92.6&63.9 &81.8&86.2&79.6\\

&XCon~\cite{fei2022xcon}&52.1&54.3&51.0 &40.5&58.8&31.7 &47.7&44.4&49.4 &96.0&97.3&95.4 &74.2&81.2&60.3 &77.6&93.5&69.7\\
&OpenCon~\cite{sun2022opencon}&54.7&63.8&54.7 &49.1&78.6&32.7 &-&-&- &-&-&- &-&-&- &84.0&93.8&81.2\\
&PromptCAL~\cite{zhang2023promptcal}&62.9&64.4&62.1 &50.2&70.1&40.6 &52.2&52.2&52.3 &\bf{97.9}&96.6&\underline{98.5} &81.2&84.2&75.3 &83.1&92.7&78.3\\
&DCCL~\cite{pu2023dynamic}&63.5&60.8&64.9 &43.1&55.7&36.2 &-&-&- &96.3&96.5&96.9 &75.3&76.8&70.2 &80.5&90.5&76.2\\
&GPC~\cite{Zhao_2023_ICCV}&52.0&55.5&47.5 &38.2&58.9&27.4 &43.3&40.7&44.8 &90.6&97.6&87.0 &75.4&{84.6}&60.1 &75.3&93.4&66.7\\

&PIM~\cite{chiaroni2023parametric}&62.7&75.7&56.2 &43.1&66.9&31.6 &-&-&- &94.7&97.4&93.3 &78.3&84.2&66.5 &83.1&\underline{95.3}&77.0\\
&$\mu$GCD~\cite{vaze2023no} &65.7&68.0&64.6 &56.5&68.1&50.9 &53.8&55.4&53.0 &-&-&- &-&-&- &-&-&-\\
&InfoSieve~\cite{rastegar2023learn}&69.4&\bf{77.9}&65.2 &55.7&74.8&46.4 &56.3&63.7&52.5 &94.8&97.7&93.4 &78.3&82.2&70.5 &80.5&93.8&73.8\\
&CiPR~\cite{hao2023cipr} &57.1&58.7&55.6 &47.0&61.5&40.1 &-&-&- 
&\underline{97.7}&97.5&97.7 &{81.5}&82.4&{79.7} &80.5&84.9&78.3\\
&SPTNet~\cite{wang2024sptnet} &65.8&68.8&65.1 &59.0&\underline{79.2}&49.3 &\underline{59.3}&61.8&58.1
&{97.3}&95.0&\bf{98.6} &81.3&84.3&75.6 &{85.4}&93.2&{81.4}\\
&CMS~\cite{choi2024contrastive}&68.2&\underline{76.5}&64.0 &56.9&76.1&47.6 &56.0&63.4&52.3 
&-&-&- &\underline{82.3}&\bf{85.7}&75.5 &84.7&\bf{95.6}&79.2\\
&AMEND~\cite{Banerjee_2024_WACV} &64.9&75.6&59.6 &52.8&61.8&48.3 &56.4&\bf{73.3}&48.2 &96.8&94.6&97.8 &81.0&79.9&\bf{83.3} &83.2&92.9&78.3\\

\cmidrule{2-20}
&GCD~\cite{vaze2022generalized}&51.3&56.6&48.7 &39.0&57.6&29.9 &45.0&41.1&46.9 &91.5&{97.9}&88.2 &73.0&76.2&66.5 &74.1&89.8&66.3\\
&\cellcolor{my_blue}\bf{Hyp-GCD}&\cellcolor{my_blue}61.0&\cellcolor{my_blue}67.0&\cellcolor{my_blue}58.0 &\cellcolor{my_blue}50.8&\cellcolor{my_blue}60.9&\cellcolor{my_blue}45.8 &\cellcolor{my_blue}48.2&\cellcolor{my_blue}43.6&\cellcolor{my_blue}50.5 
&\cellcolor{my_blue}92.9&\cellcolor{my_blue}97.5&\cellcolor{my_blue}90.6
&\cellcolor{my_blue}74.0&\cellcolor{my_blue}80.0&\cellcolor{my_blue}62.0 &\cellcolor{my_blue}80.4&\cellcolor{my_blue}92.5&\cellcolor{my_blue}74.4\\

&SimGCD~\cite{wen2023parametric}&60.3&65.6&57.7 &53.8&71.9&45.0 &54.2&59.1&51.8 &97.1&95.1&98.1 &80.1&81.2&77.8 &83.0&93.1&77.9\\
&\cellcolor{my_blue}\bf{Hyp-SimGCD}&\cellcolor{my_blue}64.8&\cellcolor{my_blue}65.8&\cellcolor{my_blue}64.2 &\cellcolor{my_blue}\underline{62.8}&\cellcolor{my_blue}73.4&\cellcolor{my_blue}\textbf{57.7} &\cellcolor{my_blue}58.7&\cellcolor{my_blue}58.9&\cellcolor{my_blue}\underline{58.5} 
&\cellcolor{my_blue}96.8&\cellcolor{my_blue}95.9&\cellcolor{my_blue}97.2 &\cellcolor{my_blue}\bf{82.4}&\cellcolor{my_blue}83.1&\cellcolor{my_blue}\underline{81.2} &\cellcolor{my_blue}\underline{86.5}&\cellcolor{my_blue}93.7&\cellcolor{my_blue}\bf{83.0}\\
&SelEx~\cite{RastegarECCV2024}&\underline{73.6}&75.3&\underline{72.8} &58.5&75.6&50.3 &57.1&64.7&53.3 
&95.9&\underline{98.1}&94.8 &\underline{82.3}&\underline{85.3}&76.3 &83.1&93.6&77.8\\
&\cellcolor{my_blue}\bf{Hyp-SelEx}&\cellcolor{my_blue}\bf{79.8}&\cellcolor{my_blue}75.8&\cellcolor{my_blue}\bf{81.8} &\cellcolor{my_blue}\bf{62.9}&\cellcolor{my_blue}\bf{80.0}&\cellcolor{my_blue}\underline{54.7} &\cellcolor{my_blue}\bf{65.9}&\cellcolor{my_blue}\underline{67.3}&\cellcolor{my_blue}\bf{65.1}
&\cellcolor{my_blue}96.7&\cellcolor{my_blue}97.6&\cellcolor{my_blue}96.3 
&\cellcolor{my_blue}\bf{82.4}&\cellcolor{my_blue}85.1&\cellcolor{my_blue}77.0
&\cellcolor{my_blue}\bf{86.8}&\cellcolor{my_blue}94.6&\cellcolor{my_blue}\underline{82.8}\\
\midrule

\multirow{9}{*}{\rotatebox{90}{\emph{DINOv2}}}
&$\mu$GCD~\cite{vaze2023no}&74.0&75.9&73.1 &76.1&91.0&68.9 &66.3&68.7&65.1 &-&-&- &-&-&- &-&-&-\\
&CiPR~\cite{hao2023cipr}&78.3&73.4&80.8 &66.7&77.0&61.8 &-&-&- &\bf{99.0}&{98.7}&99.2 &\underline{90.3}&89.0&\underline{93.1} &88.2&87.6&{88.5}\\
&SPTNet~\cite{wang2024sptnet} &76.3&79.5&74.6 &-&-&- &-&-&- &-&-&- &-&-&- &{90.1}&{96.1}&87.1\\
\cmidrule{2-20}
&GCD~\cite{vaze2022generalized}&71.9&71.2&72.3 &65.7&67.8&64.7 &55.4&47.9&59.2 &97.8&\bf{99.0}&97.1 &79.6&84.5&69.9 &78.5&89.5&73.0\\
&\cellcolor{my_blue}\bf{Hyp-GCD}&\cellcolor{my_blue}75.6&\cellcolor{my_blue}75.1&\cellcolor{my_blue}75.9 &\cellcolor{my_blue}72.8&\cellcolor{my_blue}80.4&\cellcolor{my_blue}69.1 &\cellcolor{my_blue}62.7&\cellcolor{my_blue}70.0&\cellcolor{my_blue}59.0
&\cellcolor{my_blue}97.5&\cellcolor{my_blue}\underline{98.9}&\cellcolor{my_blue}96.8 &\cellcolor{my_blue}84.5&\cellcolor{my_blue}87.5&\cellcolor{my_blue}78.5 &\cellcolor{my_blue}82.9&\cellcolor{my_blue}92.4&\cellcolor{my_blue}78.2\\

&SimGCD~\cite{wen2023parametric}&71.5&78.1&68.3 &71.5&81.9&66.6 &63.9&69.9&60.9 &98.7&96.7&\bf{99.7} &88.5&{89.2}&{87.2} &89.9&95.5&87.1\\
&\cellcolor{my_blue}\bf{Hyp-SimGCD}&\cellcolor{my_blue}77.6&\cellcolor{my_blue}77.9&\cellcolor{my_blue}77.4 &\cellcolor{my_blue}\underline{82.5}&\cellcolor{my_blue}85.8&\cellcolor{my_blue}\bf{81.0} &\cellcolor{my_blue}76.4&\cellcolor{my_blue}70.3&\cellcolor{my_blue}\underline{79.4} 
&\cellcolor{my_blue}\underline{98.9}&\cellcolor{my_blue}97.7&\cellcolor{my_blue}\underline{99.5} &\cellcolor{my_blue}\bf{91.5}&\cellcolor{my_blue}90.0&\cellcolor{my_blue}\bf{94.6} &\cellcolor{my_blue}\underline{91.9}&\cellcolor{my_blue}\underline{96.2}&\cellcolor{my_blue}\underline{89.8}\\
&SelEx~\cite{RastegarECCV2024} &\underline{87.4}&\underline{85.1}&\underline{88.5} &82.2&\bf{93.7}&76.7 &\underline{79.8}&\bf{82.3}&78.6 
&98.5*&98.8*&98.5* &87.7*&\underline{90.8}*&81.5* &90.9*&\underline{96.2}*&88.3*\\
&\cellcolor{my_blue}\bf{Hyp-SelEx}&\cellcolor{my_blue}\bf{90.7}&\cellcolor{my_blue}\bf{85.3}&\cellcolor{my_blue}\bf{93.4} &\cellcolor{my_blue}\bf{83.8}&\cellcolor{my_blue}\underline{93.3}&\cellcolor{my_blue}\underline{79.2} &\cellcolor{my_blue}\bf{83.4}&\cellcolor{my_blue}\underline{82.0}&\cellcolor{my_blue}\bf{84.1} 
&\cellcolor{my_blue}98.6&\cellcolor{my_blue}98.1&\cellcolor{my_blue}98.9 
&\cellcolor{my_blue}88.6&\cellcolor{my_blue}\bf{91.5}&\cellcolor{my_blue}82.8
&\cellcolor{my_blue}\bf{92.3}&\cellcolor{my_blue}\bf{96.4}&\cellcolor{my_blue}\bf{90.2}\\
\bottomrule
\end{tabular}
}
}
\RaggedRight
\noindent\footnotesize{*results from our implementation.}
\label{tab:ssb}
\end{table*}

\subsection{Setups and Implementations}
\noindent\textbf{Datasets.}
We thoroughly evaluate our method across diverse benchmarks, including the generic image recognition datasets CIFAR-10 and CIFAR-100~\cite{krizhevsky2009learning}, as well as ImageNet-100~\cite{deng2009imagenet}. Additionally, we assess our approach on the Semantic Shift Benchmark (SSB)~\cite{vaze2022semantic}, which includes fine-grained datasets such as CUB~\cite{wah2011caltech}, Stanford-Cars~\cite{krause20133d}, and FGVC-Aircraft~\cite{maji2013fine}. 
For each dataset, we adhere to the data split scheme detailed in~\cite{vaze2022generalized}. The method involves sampling a subset of all classes as the known (`Old') classes $\mathbf{Y}_l$. Subsequently, $50\%$ of the images from these known classes are utilized to construct $\mathbf{D}_l$, while the remaining images are designated as the unlabelled data $\mathbf{D}_u$.

\noindent\textbf{Evaluation Metrics.}
We evaluate the performance using the clustering accuracy ($\textit{ACC}$) as defined in the literature~\cite{vaze2022generalized}. The $\textit{ACC}$ on $\mathbf{D}_u$ is computed as follows, given the ground truth $y_i$ and the predicted labels $\hat{y}_i$:~$
\textstyle
    \textit{ACC}=\frac{1}{|\mathbf{D}_u|}\sum\nolimits_{i=1}^{|\mathbf{D}_u|}\mathbbm{1}(y_i=h(\hat{y}_i))$,
where $h$ denotes the optimal permutation that aligns the predicted cluster assignments with the ground-truth labels. 
The $\textit{ACC}$ values for the `All', `Old' and `New' classes are reported separately.

\noindent\textbf{Implementation Details.}
We evaluate \textit{HypCD} against the non-parametric baseline GCD~\cite{vaze2022generalized}, the parametric baseline SimGCD~\cite{wen2023parametric}, and the SOTA method SelEx~\cite{RastegarECCV2024}, utilizing both DINO~\cite{caron2021emerging} and DINOv2~\cite{oquab2023dinov2} pretrained weights. Detailed information regarding SelEx can be found in the supplementary materials. 
For GCD~\cite{vaze2022generalized}, the output dimension of the projection head $\rho_r$ is $256$. In the case of SimGCD~\cite{wen2023parametric}, the feature dimension from backbone $\phi$ is $768$. 
$\rho_r$ and the final block of $\phi$ are optimized using the SGD optimizer, with an initial learning rate of $0.1$, which is decayed to $0.001$ over time according to a cosine annealing schedule. 
The \texttt{HypFFN} is optimized using the Riemannian Adam optimizer~\cite{becigneulriemannian}, with a constant learning rate of $0.01$. All models are trained for $200$ epochs using a batch size of $128$. 
The curvature parameter $c$ is set to $0.05$ for the fine-grained datasets and $0.01$ for the generic datasets. 
Following baselines, the balancing factor $\lambda_{b}^{\mathbb{H}}$ is set to $0.35$. By default, the loss weight $\alpha_d$ increases linearly from $0$ to $1.0$.

\subsection{Quantitative Comparison}
We compare our method with recent GCD methods (including ORCA~\cite{cao2022open}, GCD~\cite{vaze2022generalized}, XCon~\cite{fei2022xcon}, OpenCon~\cite{sun2022opencon},  PromptCAL~\cite{zhang2023promptcal}, DCCL~\cite{pu2023dynamic}, GPC~\cite{Zhao_2023_ICCV}, CiPR~\cite{hao2023cipr}, SimGCD~\cite{wen2023parametric}, $\mu$GCD~\cite{vaze2023no}, InfoSieve~\cite{rastegar2023learn}, SPTNet~\cite{wang2024sptnet}, CMS~\cite{choi2024contrastive}, AMEND~\cite{Banerjee_2024_WACV} and SelEx~\cite{RastegarECCV2024}) and report the results in Tab.~\ref{tab:results}. 
The evaluation encompasses performance on the SSB benchmark~\cite{vaze2022semantic} and generic datasets~\cite{krizhevsky2009learning, deng2009imagenet}. 
The hyperbolic methods applying our \textit{HypCD} framework are indicated by the `Hyp-' prefix.

\begin{figure}[htb]    
    \centering
    \includegraphics[width=0.47\textwidth]{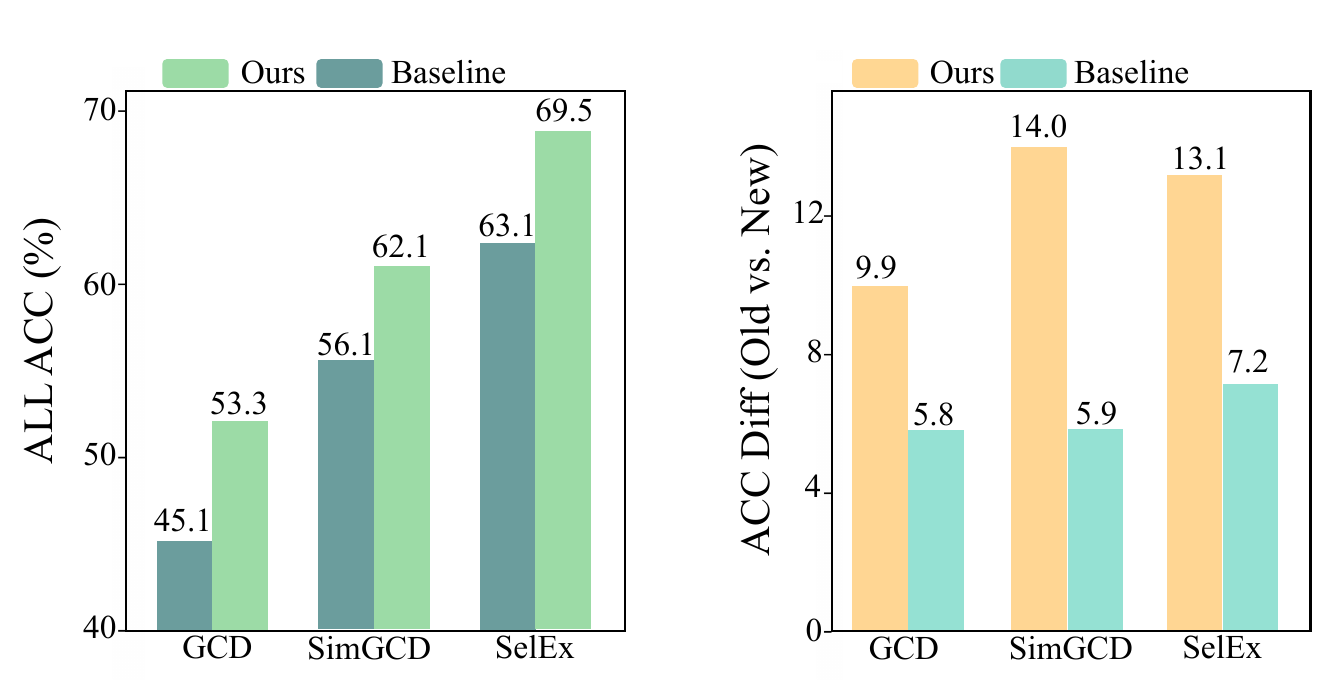}
    \caption{Comparison of baseline and hyperbolic counterparts on the SSB. Left: `All' \textit{ACC} (higher is better). Right: Discrepancy between `Old' and `New' \textit{ACC} (smaller is better). }
    \label{fig:results}
\end{figure}

\noindent\textbf{Results on SSB.}
The performance of the GCD methods on the SSB benchmark, utilizing both DINO~\cite{caron2021emerging} and DINOv2~\cite{oquab2023dinov2} pretrained weights, is summarized in the left section of Tab.~\ref{tab:results}. 
Besides, we provide a comparative analysis between the three baseline methods and their hyperbolic counterparts with DINO backbone in Fig.~\ref{fig:results}. 
Our hyperbolic methods consistently outperform their Euclidean counterparts, with particularly strong results observed when utilizing the DINOv2 backbone. 
Among the evaluated methods, Hyp-SelEx achieves the highest average accuracy (\textit{ACC}) across all datasets, notably excelling on the CUB dataset, where it records an accuracy of 79.8\% for the `All' classes with DINO and 90.7\% with DINOv2, establishing it as the leading approach. 
More strikingly, on the Stanford-Cars dataset, our Hyp-GCD method outperforms the baseline by 11.8\%, 13.3\% and 15.9\% in terms of \textit{ACC} for the `All', `Old' and `New' categories, respectively. 
Fig.~\ref{fig:results} (left) illustrates the average \textit{ACC} on `All' categories across the three datasets in the SSB benchmark, indicating that our hyperbolic methods surpass the baseline by a margin of at least 6.0\%. 
Furthermore, as shown in Fig.~\ref{fig:results} (right), hyperbolic methods exhibit a consistently smaller \textit{ACC} gap between `Old' and `New' classes, highlighting the effectiveness of \textit{HypCD} in enhancing knowledge transfer from known to unseen categories. Additionally, DINOv2 outperforms DINO across all methods, underscoring its ability to capture complex data representations more effectively.

\noindent\textbf{Results on Generic Datasets.} 
In the right section of Tab.~\ref{tab:results}, we present the results on three widely used generic datasets: CIFAR-10~\cite{krizhevsky2009learning}, CIFAR-100~\cite{krizhevsky2009learning}, and ImageNet-100~\cite{deng2009imagenet}. 
Our methods demonstrate consistent improvements across all cases, regardless of the backbone employed. Notably, these enhancements are especially significant on CIFAR-100 and ImageNet-100, which present greater challenges compared to CIFAR-10, where performance is nearly saturated. 
For CIFAR-100, Hyp-SimGCD and Hyp-SelEx achieve the highest accuracy of 82.4\% for the `All' categories using DINO, while Hyp-SimGCD ranks first with an accuracy of 91.5\% on this metric when utilizing DINOv2, significantly surpassing baseline methods and the previous SOTA. 
Results on ImageNet-100 further validate the effectiveness of hyperbolic methods; Hyp-SelEx achieves the highest performance across `All', `Old', and `New' categories with both DINO and DINOv2, outperforming the baseline by a margin of up to 3.7\%.

\begin{table}[ht]
\centering
\caption{Experimental results using different $c$, $r$ and $\alpha_d^{\text{max}}$ values in Hyp-SimGCD with DINO~\cite{caron2021emerging} pre-trained backbone. Results on the CUB~\cite{wah2011caltech} and CIFAR-100~\cite{krizhevsky2009learning} datasets are reported.}
\setlength{\tabcolsep}{3mm}{
\resizebox{0.45\textwidth}{!}{
\begin{tabular}{lcccccc}
        \toprule
        &\multicolumn{3}{c}{Stanford-Cars~\cite{krause20133d}}&\multicolumn{3}{c}{CIFAR-100~\cite{krizhevsky2009learning}}\\
        \cmidrule(lr{1em}){2-4} \cmidrule(lr{1em}){5-7}
        parameter&All&Old&New&All&Old&New\\
        \midrule
        ~~~$c$~~=~~0.01&61.4&74.4&55.1 &\bf{82.4}&83.1&\bf{81.2}\\
        ~~~$c$~~=~~0.05&\bf{62.8}&73.4&\bf{57.7} &81.6&\bf{84.0}&76.7\\
        ~~~$c$~~=~~0.1&62.3&\bf{75.1}&56.1 &81.1&82.3&78.8\\
        % $c$~~=~~0.3&&& &&&\\
        \midrule
        ~~~$r$~~=~~1.0&60.0&72.9&53.7 &\bf{82.4}&\bf{83.1}&\bf{81.2}\\
        ~~~$r$~~=~~1.5&61.2&\bf{75.7}&54.2 &81.2&82.4&78.8\\
        ~~~$r$~~=~~2.3&\bf{62.8}&73.4&\bf{57.7} &80.1&81.1&78.3\\
        \midrule
        $\alpha_d^{\text{max}}$=~~0.1&59.6&\bf{77.5}&51.0 &81.3&82.4&79.1\\
        $\alpha_d^{\text{max}}$=~~0.5&62.0&77.2&54.6 &\bf{82.4}&83.1&\bf{81.2}\\
        $\alpha_d^{\text{max}}$=~~1.0&\bf{62.8}&73.4&\bf{57.7} &78.9&\bf{83.5}&69.7\\
        \bottomrule
        \end{tabular}
}
}
\label{tab:curv}
\end{table}

\subsection{Impact of Hyperparameters}
\noindent\textbf{Manifold Curvature.}
Building on previous studies~\cite{Khrulkov_2020_CVPR, atigh2022hyperbolic} that explore the application of hyperbolic geometry across various tasks, the curvature parameter $c$ (as discussed in Sec.~\ref{sec:method:hyp_space}) is a crucial factor influencing performance and may yield different optimal values across datasets and methods. 
Intuitively, as the value of $c$ approaches $0$, the radius tends toward infinity, causing the Poincaré ball to flatten and resemble Euclidean space; conversely, larger values of $c$ correspond to a steeper configuration. 
The widely accepted range for $c$ is between $0.01$ and $0.3$~\cite{ermolov2022hyperbolicvisiontransformerscombining}, with larger values exceeding this range resulting in performance degradation. 
In our experiments, we evaluate different curvature values of $0.01$, $0.05$, and $0.1$ using Hyp-SimGCD, as presented in Tab.~\ref{tab:curv}. 
Our findings indicate that the optimal curvature values differ between generic and fine-grained datasets. For fine-grained datasets such as CUB~\cite{wah2011caltech}, the optimal value is $0.05$, while for generic datasets like CIFAR-100, a value of $0.01$ proves to be more effective.

\noindent\textbf{Clipping Value.}
As articulated in~\cite{guo2022clippedhyperbolicclassifierssuperhyperbolic}, feature clipping has emerged as a standard technique for training hyperbolic neural networks. 
In our framework, we also observe that it plays a crucial role in category discovery performance. 
In line with the methodology outlined in~\cite{guo2022clippedhyperbolicclassifierssuperhyperbolic}, we investigate a range of clipping values, specifically $1.0$, $1.5$, and $2.3$. 
The results shown in the second row of Tab.~\ref{tab:curv} demonstrate that optimal clipping values vary between fine-grained and generic datasets. 
For fine-grained datasets like CUB~\cite{wah2011caltech}, the optimal clipping value is determined to be $2.3$. 
Conversely, for generic datasets like CIFAR-100~\cite{krizhevsky2009learning}, a clipping value of $1.0$ is shown to be more effective.

\noindent\textbf{Loss Weight.} 
As detailed in Sec.~\ref{sec:method:hyp_cd}, we implement a hybrid contrastive loss that combines both distance-based and angle-based components, which is essential for effective optimization in hyperbolic space. A loss weight, denoted as $\alpha_d$, is introduced to regulate the balance between these two types of losses and linearly increasing from $0$. 
In the initial stages, the model prioritizes optimizing the angle between sample points and progressively shifts focus toward optimizing the hyperbolic distance. 
Consistent with the observations for curvature and clipping values, the optimal max value $\alpha_d^{\text{max}}$, varies considerably between coarse-grained and fine-grained datasets. 
For fine-grained datasets such as CUB~\cite{wah2011caltech}, an optimal value of $1.0$ is observed, whereas for more generic datasets like CIFAR-100~\cite{krizhevsky2009learning}, a value of $0.5$ is found to yield better performance.

\subsection{Qualitative Comparison}
\begin{figure}[t]    
    \centering
    \includegraphics[width=0.47\textwidth]{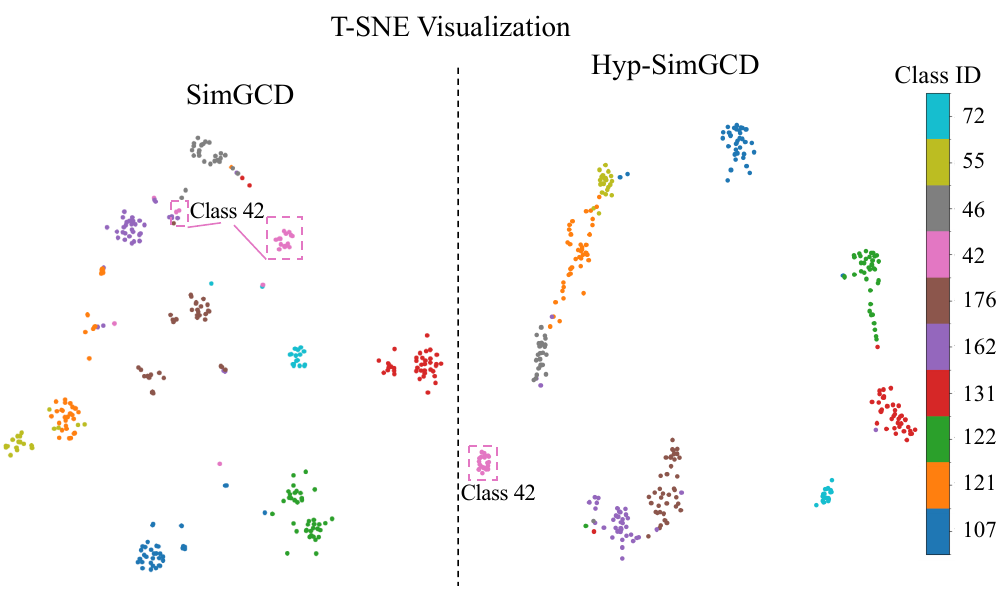}
    \caption{T-SNE~\cite{van2008visualizing} comparison between SimGCD~\cite{wen2023parametric} and our Hyp-SimGCD using 40 randomly sampled instances from 10 randomly selected categories of the Stanford-Cars dataset~\cite{krause20133d}.}
    \label{fig:tsne}
\end{figure}
In Fig.~\ref{fig:tsne}, we present a t-SNE~\cite{van2008visualizing} visualization of features extracted from the backbone, represented as $\mathbf{z}_i = \phi(\mathbf{x}_i)$. This visualization compares SimGCD with our Hyp-SimGCD. 
On the left side of the figure, the clusters generated by SimGCD appear dispersed. Data points from Class 42, highlighted in pink, are spread across multiple areas, indicating significant overlap and a lack of compactness. 
In contrast, Hyp-SimGCD creates more distinct and tightly clustered groups, concentrating the data points of Class 42 in a more confined area.  
This comparison implies that Hyp-SimGCD enhances both intra-class compactness and inter-class separation through our hyperbolic representation and classifier learning method. 
Importantly, even within the original Euclidean space of the backbone network, Hyp-SimGCD exhibits robust clustering performance, which arises from the properties of hyperbolic space in encoding hierarchical structures.

\section{Conclusion}
\label{sec:conclusion}
In this paper, we investigate a previously overlooked perspective in GCD by utilizing a representation space that captures the hierarchical structure of each sample, instead of the conventional Euclidean or spherical spaces. 
Our approach leverages the distinctive properties of hyperbolic space, where the volume increases exponentially with radius. This characteristic makes hyperbolic space especially suitable for modelling data possessing hierarchical structures, thereby enhancing representational capacity for category discovery. 
We propose a simple yet effective framework, \textit{HypCD}, for integrating hyperbolic geometry into GCD methods. Through extensive experiments with parametric and non-parametric GCD baselines and the SOTA method, our framework consistently demonstrates superior performance on public benchmarks, underscoring the effectiveness of hyperbolic space for category discovery.

\paragraph{Acknowledgements.} 
This work is supported by National Natural Science Foundation of China (Grant No. 62306251),  Hong Kong Research Grant Council - Early Career Scheme (Grant No. 27208022) and General Research Fund (Grant No. 17211024), and HKU Seed Fund for Basic Research.

{
    \small
    \bibliographystyle{ieeenat_fullname}
    \bibliography{main}
}

\maketitlesupplementary

We provide additional details and experimental results in this supplementary material, which is organized as follows:
\begin{itemize}
\item \S\ref{sec:details} More Experimental Details
\item \S\ref{sec:quantitative} More Quantitative Results
\item \S\ref{sec:qualitative} More Qualitative Results

\end{itemize}

\section{More Experimental Details}
\label{sec:details}
\subsection{Dataset Statistics}
For each dataset, we adhere to the data splitting scheme described in~\cite{vaze2022generalized}. In this scheme, $50\%$ of the classes will be sampled as `Old', with the exception of CIFAR-100, which samples $80\%$ of the classes. Following this, $50\%$ of the images from known classes are used to create the labelled dataset $\mathbf{D}_l$, while the remaining images are allocated to the unlabelled dataset $\mathbf{D}_u$. 
The statistics for all the datasets utilized in this work are summarized in Tab.~\ref{tab:datasets}.

\begin{table}[h!]
\centering
\caption{Overview of the dataset, including the classes in the labelled and unlabelled sets ($M=|\mathbf{Y}_l|$, $K=|\mathbf{Y}_l \cup \mathbf{Y}_u|$) and counts of images ($|\mathbf{D}_l|$, $|\mathbf{D}_u|$). `FG' denotes fine-grained.}
\setlength{\tabcolsep}{3mm}{
\resizebox{0.45\textwidth}{!}{
\begin{tabular}{lccccc}
    \toprule
    Dataset &FG &$|\mathbf{D}_l|$&$M$&$|\mathbf{D}_u|$ &$K$\\
    \midrule
    CIFAR-10~\cite{krizhevsky2009learning} &\ding{55} & 12.5K & 5 & 37.5K & 10 \\
    CIFAR-100~\cite{krizhevsky2009learning} &\ding{55} & 20.0K & 80 & 30.0K & 100 \\
    ImageNet-100~\cite{deng2009imagenet} &\ding{55} & 31.9K & 50 & 95.3K & 100 \\
    CUB~\cite{wah2011caltech} &\ding{51} & 1.5K & 100 & 4.5K & 200 \\
    Stanford-Cars~\cite{krause20133d} &\ding{51} & 2.0K & 98 & 6.1K & 196 \\
    FGVC-Aircraft~\cite{maji2013fine} & \ding{51} &1.7K & 50 & 5.0K & 100 \\
    Herbarium19~\cite{tan2019herbarium} &\ding{51} & 8.9K & 341 & 25.4K & 683 \\
    Oxford-Pet~\cite{parkhi2012cats}    & \ding{51} &0.9K &19 &2.7K &37 \\
    \bottomrule
\end{tabular}
}
}
\label{tab:datasets}
\end{table}

\subsection{Additional Implementation Details}
Consistent with prior studies~\cite{RastegarECCV2024, wen2023parametric, vaze2022generalized}, we employ the ViT-B architecture~\cite{dosovitskiy2020image} with pretrained weights from either DINO~\cite{caron2021emerging} or DINOv2~\cite{oquab2023dinov2} as our backbone network.  
For our proposed hyperbolic methods, we adhere to nearly all hyperparameter settings established in \cite{RastegarECCV2024, wen2023parametric, vaze2022generalized} to facilitate fair comparisons with their respective baselines. The specific details are summarized as follows: 
For Hyp-SimGCD and Hyp-GCD, only the last block of the backbone is fine-tuned across all datasets. 
In contrast, Hyp-SelEx implements dataset-specific fine-tuning: the last two blocks are fine-tuned for CUB~\cite{wah2011caltech}, FGVC-Aircraft~\cite{maji2013fine}, and all generic datasets, while the last three blocks are fine-tuned for Stanford-Cars~\cite{krause20133d}.  
Regarding method-specific hyperparameters, for Hyp-SimGCD, we set the weight $\xi$, which controls the weight of mean entropy loss, to $1.0$ for all the datasets. 
For Hyp-SelEx, we follow~\cite{RastegarECCV2024} in setting $\alpha$, which regulates label smoothing, to $0.5$ for FGVC-Aircraft~\cite{maji2013fine}, $1.0$ for CUB~\cite{wah2011caltech} and Stanford-Cars~\cite{krause20133d}, and $0.1$ for generic datasets. 
Additionally, the proposed parameter $\alpha_d$, which balances distance-based and angle-based losses, linearly increases from $0$ to its maximum value during training according to the formula: $\alpha_{d} = \frac{{e}*\alpha_{d}^{\text{max}}}{200}$, where $e$ is the current training epoch. 
Specifically, we set $\alpha_{d}^{\text{max}}$ to $1$ for fine-grained and $0.5$ for generic datasets. 

\subsection{Details of Hyp-SelEx}
\label{sec:selex}
\cite{RastegarECCV2024} proposes a hierarchical non-parametric method, SelEx, to address fine-grained GCD through a novel concept of \textit{self-expertise}. 
It begins by constructing hierarchical pseudo-labeling via a \textit{balanced semi-supervised k-means} algorithm to initialize clusters for known categories and then iteratively refines them by incorporating an equal number of random samples for unseen categories to balance cluster distribution. 
Following it, \textit{supervised self-expertise} leverages weakly-supervised pseudo labels to group samples by capturing abstract-level similarity, whereas \textit{unsupervised self-expertise} focuses on distinguishing semantically similar hard negative samples within the same clusters to sharpen fine-grained categorization. 

Its representation learning objective composes of unsupervised self-expertise loss $\mathcal{L}_{\text{\text{USE}}}$ and supervised self-expertise loss $\mathcal{L}_{\text{SSE}}$. 
The unsupervised self-expertise loss, defined as $\mathcal{L}_{\text{USE}} = \ell_{ce} (\mathbf{p}, \hat{\mathbf{t}})$, calculates the binary cross entropy loss between the logits $\mathbf{p}$ and an adjusted target $\hat{\mathbf{t}}$, where $\mathbf{p}$ is calculated based on Euclidean distance, unlike prior GCD~\cite{vaze2022generalized} approach that utilizes cosine similarity. 
\cite{RastegarECCV2024} introduces an adjusted target matrix $\mathbf{\hat{t}}$ to recalibrate targets based on semantic similarity between samples. 
Specifically, $\mathbf{\hat{t}} = \alpha \mathbf{t} + (1 - \alpha) \mathbf{I}$, where $\mathbf{t}$ can be calculated using $\mathbf{t} = [\sum_{k=1}^{\lg K} \frac{\mathbbm{1}(\hat{y}_i^k \neq \hat{y}_j^k)}{2^k}]$ based on pseudo label $\hat{y}_i^k$ and $\hat{y}_j^k$ from hierarchical level $k$. $\alpha$ is the hyperparameter to control the label smoothing by identity metric $\mathbf{I}$. 
Then, the hierarchical supervised self-expertise loss can be denoted as:
\begin{equation}
    \mathcal{L}_{\text{SSE}} = \frac{1}{2} \left( \sum_{k=0}^{\lg K} \frac{\mathcal{L}_s^k | \frac{\mathbf{d}}{2^k}}{2^k} \right),
\end{equation}
where $\mathcal{L}_s^k | \frac{\mathbf{d}}{2^k}$ represents the supervised representation loss applied exclusively to the segment $\frac{\mathbf{d}}{2^k}$ of the embedding vector $\mathbf{d}$, corresponding to each level of the hierarchy. 
The final representation loss is given by $\mathcal{L}_{rep}=(1-\lambda_b)\mathcal{L}_{\text{USE}} + \lambda_b \mathcal{L}_{\text{SSE}}$. 
To combine SelEx with hyperbolic embeddings, we extend the hierarchical representation learning used in \cite{RastegarECCV2024} into the hyperbolic space, utilizing the methodology introduced in the main paper.

Following the above pace, our Hyp-SelEx utilizes hyperbolic supervised and unsupervised self-expertise, denoted as $\mathcal{L}_{\text{SSE}}^\mathbb{H}$ and $\mathcal{L}_{\text{USE}}^\mathbb{H}$, respectively. 
Given two randomly augmented views $\mathbf{x}_i$ and $\mathbf{x}_i'$ for the same image in a mini-batch $B$,  $\mathbf{z}_i$ and $\mathbf{z}_i'$ represent the feature extracted from backbone network $\phi$ and projector $\rho_r$ of these two views in the Euclidean space, represented as $\mathbf{z}_i = \rho_r(\phi(\mathbf{x}_i))$. 
As introduced in Sec.3.4 of the main paper, we employ a hybrid of distance-based and angle-based loss functions, and hence the unsupervised self-expertise loss is represented as:
\begin{equation}
    \mathcal{L}_{\text{USE}}^\mathbb{H} = \alpha_d \ell_{ce} (\mathbf{p}_{\text{dis}}, \mathbf{\hat{t}}) + (1- \alpha_d) \ell_{ce} (\mathbf{p}_{\text{ang}}, \mathbf{\hat{t}}),
\end{equation}
where $\mathbf{p}_{\text{dis}}$ is the logit calculated based on the negative hyperbolic distance, expressed as $\mathcal{S}_d(\mathcal{M}(\mathbf{z}_i), \mathcal{M}(\mathbf{z}_i'))$, and $\mathbf{p}_{\text{ang}}$ is the logit calculated based on the original distance metrics, expressed as $\mathcal{S}_a(\mathcal{M}(\mathbf{z}_i), \mathcal{M}(\mathbf{z}_i'))$. 
Similarly, the hyperbolic supervised self-expertise loss is defined as:
\begin{equation}
\textstyle
   \mathcal{L}_{\text{SSE}}^\mathbb{H} = \frac{1}{2} \left( \sum_{k=0}^{\lg K} \frac{\alpha_d(\mathcal{L}_{dis}^k | \frac{\mathbf{d}}{2^k}) + (1-\alpha_d)(\mathcal{L}_{ang}^k | \frac{\mathbf{d}}{2^k})}{2^k} \right), 
\end{equation}
where $\mathcal{L}_{dis}^k | \frac{\mathbf{d}}{2^k} $ and $\mathcal{L}_{ang}^k | \frac{\mathbf{d}}{2^k}$ denote the hyperbolic supervised distance-based and angle-based losses applied exclusively to the segment $\frac{\mathbf{d}}{2^k}$. 
The final training objective of Hyp-SelEx is formulated as:
\begin{equation}
     \mathcal{L}_{rep}^\mathbb{H} = (1-\lambda_{b}^\mathbb{H})\mathcal{L}_{\text{USE}}^\mathbb{H} + \lambda_{b}^\mathbb{H} \mathcal{L}_{\text{SSE}}^\mathbb{H}.
\end{equation}

\begin{table}[h]
\centering
\caption{Results with the estimated number of categories, all methods use the DINO~\cite{caron2021emerging} pretrained weights. 
}
\setlength{\tabcolsep}{3mm}{
\resizebox{0.48\textwidth}{!}{
\begin{tabular}{lcccccccccccc}
    \toprule
     &\multicolumn{3}{c}{CUB~\cite{wah2011caltech}}&\multicolumn{3}{c}{Stanford-Cars~\cite{krause20133d}}\\
    \cmidrule(lr{1em}){2-4} \cmidrule(lr{1em}){5-7}
 Method&All&Old&New&All&Old&New\\ \hline
    GCD~\cite{vaze2022generalized}&47.1&55.1&44.8 &35.0&56.0&24.8\\
    SimGCD~\cite{wen2023parametric}&61.5&66.4&59.1 &49.1&65.1&41.3\\
    $\mu$GCD~\cite{vaze2023no}&62.0&60.3&62.8 &56.3&66.8&51.1\\
    SelEx~\cite{RastegarECCV2024}&\underline{72.0}&\underline{72.3}&\underline{71.9} &58.7&\underline{75.3}&50.8\\ 
    \hline
    \textbf{Hyp-GCD}&60.2&64.6&58.0 &48.1&60.2&42.2 \\
    \textbf{Hyp-SimGCD}&64.7&66.6&63.8 &\underline{60.3}&73.5&\underline{53.9} \\
    \textbf{Hyp-SelEx}&\textbf{79.6}&\textbf{75.8}&\textbf{81.6} &\textbf{62.1}&\textbf{76.2}&\textbf{55.3} \\
    \bottomrule
\end{tabular}
}
}
\label{tab:estk1}
\end{table}

\section{More Quantitative Results}
\label{sec:quantitative}
\subsection{GCD With Unknown Category Numbers}
In line with the majority of the literature~\cite{vaze2022generalized,wen2023parametric,wang2024sptnet,RastegarECCV2024}, our primary experiments presented in the main paper utilize the ground-truth category numbers. 
This section reports results based on estimated category numbers obtained from an off-the-shelf method~\cite{vaze2022generalized}, illustrating the performance of our approach when ground-truth category numbers are unavailable. 
For the CUB dataset, we estimate $K = 231$, while for Stanford-Cars, we estimate $K = 230$. In contrast, the actual ground-truth counts are $K = 200$ and $K = 196$, respectively. 
We compare our methods with SimGCD~\cite{wen2023parametric}, $\mu$GCD~\cite{vaze2023no}, and GCD~\cite{vaze2022generalized} in Table~\ref{tab:estk1}. Despite a discrepancy of approximately 
15\% between the ground-truth and estimated category numbers for both CUB~\cite{wah2011caltech} and Stanford-Cars~\cite{krause20133d}, our hyperbolic methods exhibit only a marginal decline in performance.

\begin{table}[h]
\centering
\caption{Experimental results using different embedding dimensions on Hyp-GCD with DINO~\cite{caron2021emerging} pre-trained backbone. Results on the CUB~\cite{wah2011caltech} and Stanford-Cars~\cite{krause20133d} datasets are reported.}
\setlength{\tabcolsep}{3mm}{
\resizebox{0.45\textwidth}{!}{
\begin{tabular}{ccccccc}
        \toprule
        &\multicolumn{3}{c}{CUB~\cite{wah2011caltech}}&\multicolumn{3}{c}{Stanford-Cars~\cite{krause20133d}}\\
        \cmidrule(lr{1em}){2-4} \cmidrule(lr{1em}){5-7}
        dimension&All&Old&New&All&Old&New\\
        \midrule
        64&57.6&63.6&54.6 &47.2&56.7&42.6\\
        128&59.5&65.0&56.7 &48.2&60.0&42.5\\
        256&61.0&\textbf{67.0}&58.0 &\textbf{50.8}&\textbf{60.9}&45.8\\
        512&\textbf{61.2}&65.3&\textbf{59.1} &50.3&59.5&\textbf{45.9}\\
        \bottomrule
        \end{tabular}
}
}
\label{tab:dim}
\end{table}

\subsection{Embedding Dimension}
In our framework, the parametric method Hyp-SimGCD employs the original $768$-dimensional embeddings from the pretrained ViT-B backbone. 
For the non-parametric methods, Hyp-GCD and Hyp-SelEx, we project the features from the pretrained backbone into a new spherical space using an MLP projection network, followed by an exponential mapping into hyperbolic space. 
In the baseline methods, GCD and SelEx, the final embedding dimension is set to $65,536$. 
However, our empirical findings indicate that a significantly lower dimension can yield satisfactory performance with our hyperbolic method, Hyp-GCD. 
As shown in Tab.~\ref{tab:dim}, embeddings of $256$ dimensions yield promising results for Hyp-GCD. 
This suggests that the intrinsic properties of hyperbolic space facilitate more expressive representations at lower dimensions (\eg, $256$ or $512$), effectively capturing hierarchical structures and complex relationships among data points. 
For Hyp-SelEx, we have chosen a dimension of $8,092$, which is also significantly lower than that of the baseline methods.

\begin{table}[h]
\centering
\caption{Comparison with recent GCD methods on Herbarium19~\cite{tan2019herbarium} and Oxford-Pet~\cite{parkhi2012cats}.} 
\setlength{\tabcolsep}{3mm}{
\resizebox{0.48\textwidth}{!}{
\begin{tabular}{lcccccc}
    \toprule
     &\multicolumn{3}{c}{Oxford-Pet~\cite{parkhi2012cats}}&\multicolumn{3}{c}{Herbarium19~\cite{tan2019herbarium}}\\
    \cmidrule(lr{1em}){2-4} \cmidrule(lr{1em}){5-7} 
 Method&All&Old&New&All&Old&New\\ 
    \hline
    ORCA~\cite{cao2022open}&-&-&- &24.6&26.5&23.7 \\
    GCD~\cite{vaze2022generalized}&80.2&85.1&77.6 &35.4&51.0&27.0\\
    XCon~\cite{fei2022xcon}&86.7&91.5&84.1 &-&-&-\\
    OpenCon~\cite{sun2022opencon}&-&-&- &39.3&58.9&28.6\\
    DCCL~\cite{pu2023dynamic}&88.1&88.2&88.0 &-&-&- \\
    SimGCD~\cite{wen2023parametric}&91.7&83.6&\textbf{96.0} &44.0&58.0&36.4 \\
    $\mu$GCD~\cite{vaze2023no}&-&-&- &\textbf{45.8}&\textbf{61.9}&\textbf{37.2} \\
    InfoSieve~\cite{rastegar2023learn}&90.7&\textbf{95.2}&88.4 &40.3&59.0&30.2 \\ 
    SelEx~\cite{RastegarECCV2024} &\underline{92.5}&\underline{91.9}&92.8 &39.6&54.9&31.3 \\
    \hline
    \textbf{Hyp-GCD}&86.7&85.5&87.4 &38.6&43.1&36.2\\
    \textbf{Hyp-SimGCD}&92.2&85.7&\underline{95.7} &\underline{45.1}&\underline{60.1}&\underline{36.9}\\
    \textbf{Hyp-SelEx}&\textbf{92.7}&91.5&93.3 &40.5&49.0&36.0\\
    \bottomrule
\end{tabular}
}
}
\label{tab:add}
\end{table}

\subsection{Results on Additional Datasets}
To further evaluate the proposed method, we conduct assessments on two additional fine-grained datasets: Oxford-Pet\cite{parkhi2012cats} and Herbarium19\cite{tan2019herbarium}. 
The Oxford-Pet dataset poses a significant challenge due to its variety of cat and dog species, alongside limited data availability. 
In contrast, Herbarium19 is a botanical research dataset that encompasses a wide range of plant types, characterized by its long-tailed distribution and fine-grained categorization. 
The results of our experiments on these two datasets are summarized in Tab.~\ref{tab:add}. 
Our Hyp-SelEx method achieves the highest accuracy across all categories in the Oxford-Pet dataset. Furthermore, on Herbarium19, Hyp-SelEx secures the second-best performance on all three evaluation metrics.

\begin{figure*}[htbp]
    \centering
    \includegraphics[width=0.8\linewidth]{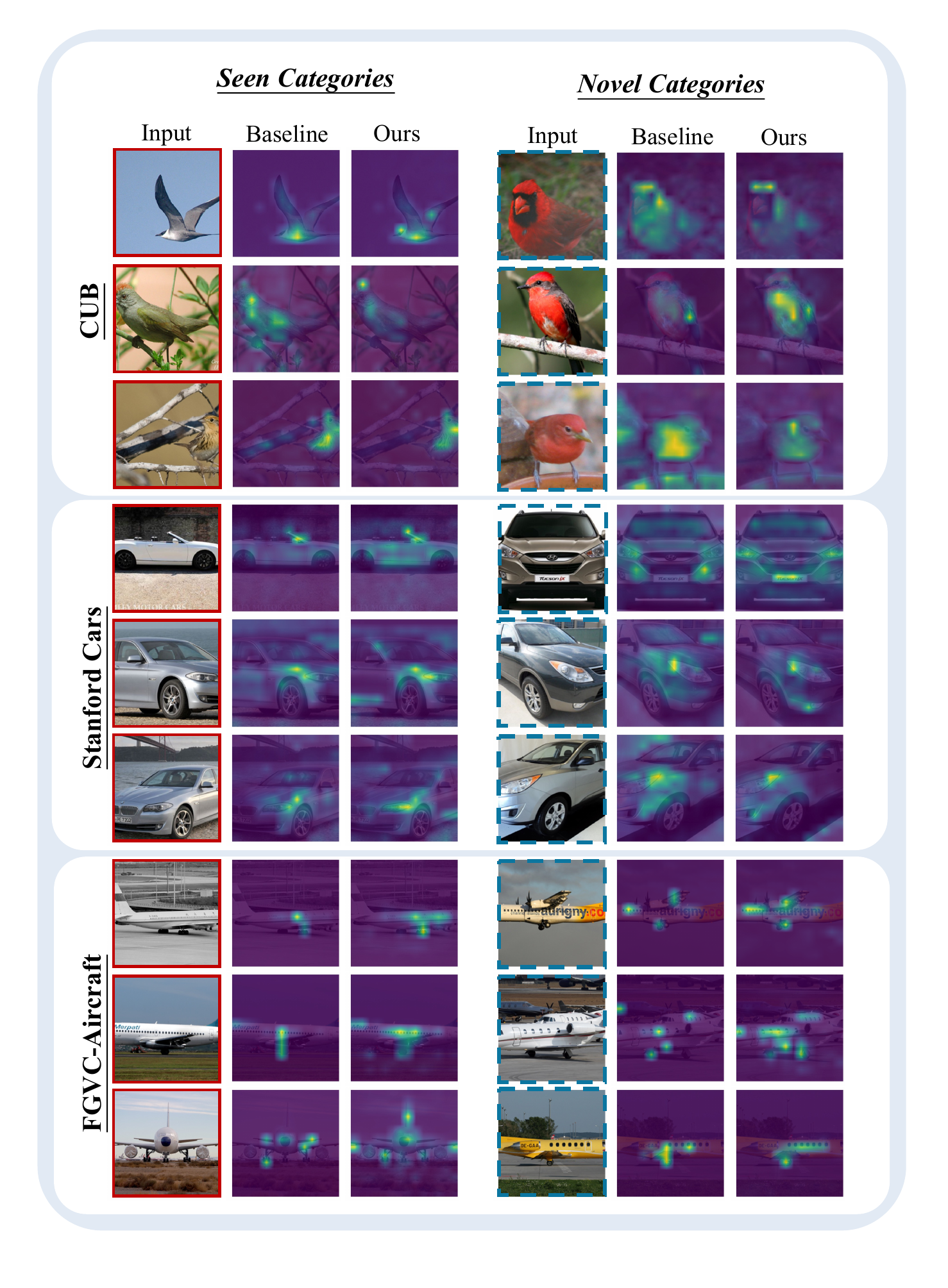}
    \caption{Visualization of attention maps of GCD~\cite{vaze2022generalized} and our Hyp-GCD.}
    \label{fig:attn_vis}
\end{figure*}

\section{More Qualitative Results}
\label{sec:qualitative}
Fig.~\ref{fig:attn_vis} displays the attention maps of GCD \cite{vaze2022generalized} and Hyp-GCD, generated from the final transformer block of the DINO backbone \cite{caron2021emerging}. These attention maps are applied across three fine-grained datasets within the SSB benchmark \cite{vaze2022semantic}. 
In this block, a \textit{multi-head self-attention} layer utilizing $12$ attention heads processes the input features, resulting in $12$ attention maps at a resolution of $14\times 14$. Following the methodology detailed in \cite{caron2021emerging}, we compute the mean value of these attention maps and subsequently upsample them to the original image resolution for visualization. 
The results indicate that our method significantly enhances focus on semantically relevant regions within the image, effectively capturing fine-grained details that are crucial for distinguishing between categories. 
In contrast, the baseline approach yields more diffuse and less targeted attention maps, often insufficiently highlighting critical areas, particularly concerning unseen categories. 
These findings emphasize the robustness and generalization capability of our method in identifying meaningful visual regions, even for novel categories, thereby demonstrating its superiority over the baseline approach.

\end{document}